# Spectrally-Encoded Single-Pixel Machine Vision Using Diffractive Networks

Jingxi Li[1,2,3,†], Deniz Mengu[1,2,3,†], Nezih T. Yardimci[1,3], Yi Luo[1,2,3], Xurong Li[1,3], Muhammed Veli[1,2,3], Yair Rivenson[1,2,3], Mona Jarrahi[1,3] and Aydogan Ozcan[1,2,3*]

[1]Electrical and Computer Engineering Department, University of California, Los Angeles, CA, 90095, USA

[2]Bioengineering Department, University of California, Los Angeles, CA, 90095, USA

[3]California NanoSystems Institute (CNSI), University of California, Los Angeles, CA, 90095, USA

[*]Correspondence to: ozcan@ucla.edu

[†]These authors contributed equally to this work.

**Abstract:**

3D engineering of matter has opened up new avenues for designing systems that can perform various computational tasks through light-matter interaction. Here, we demonstrate the design of optical networks in the form of multiple diffractive layers that are trained using deep learning to transform and encode the spatial information of objects into the power spectrum of the diffracted light, which are used to perform optical classification of objects with a single-pixel spectroscopic detector. Using a time-domain spectroscopy setup with a plasmonic nanoantenna-based detector, we experimentally validated this machine vision framework at terahertz spectrum to optically classify the images of handwritten digits by detecting the spectral power of the diffracted light at ten distinct wavelengths, each representing one class/digit. We also report the coupling of this spectral encoding achieved through a diffractive optical network with a shallow electronic neural network, separately trained to reconstruct the images of handwritten digits based on solely the spectral information encoded in these ten distinct wavelengths within the diffracted light. These reconstructed images demonstrate task-specific image decompression and can also be cycled back as new inputs to the same diffractive network to improve its optical object classification. This unique machine vision framework merges the power of deep learning with the spatial and spectral processing capabilities of diffractive networks, and can also be extended to other spectral-domain measurement systems to enable new 3D imaging and sensing modalities integrated with spectrally encoded classification tasks performed through diffractive optical networks.

## Main

Engineering of materials and material properties has opened a myriad of new opportunities for designing new components and devices with unique functionalities that were not possible before (*1–10*). Precise control of light-matter interaction at different scales has been the key behind the success of these engineered material systems, including e.g., plasmonics, metamaterials and photonic crystals, which led to various new capabilities for nanoscopic imaging and sensing as well as light generation, modulation and detection (*11–20*). This quest to harness engineered light-matter interactions has also led to all-optical processors that perform a desired computational task through wave propagation within specially designed materials (*21–23*). All the way from solving equations to performing statistical inference and machine learning, these approaches highlight the emergence of engineered and trained matter as a building block of optical computation. Considering the rapid advances being made in e.g., autonomous vehicles, robotic systems and medical imaging, there is a growing need for performing computation optically, in order to benefit from the low-power, low-latency and scalability that passive optical systems can offer.

Here we report deep learning-based design of diffractive networks that perform machine vision and statistical inference by encoding the spatial information of objects into optical spectrum through learnable diffractive layers that collectively process the information contained at multiple wavelengths to perform optical classification of objects using a single-pixel detector (Fig. 1a). Unlike conventional optical components used in machine vision systems, we employ diffractive layers that are composed of two-dimensional arrays of passive pixels, where the complex-valued transmission or reflection coefficients of individual pixels are independent learnable parameters that are optimized using a computer through deep learning and error back-propagation (*24*). The use of deep learning in optical information processing systems has emerged in various exciting directions including integrated photonics solutions (*25–32*) as well as free-space optical platforms (*33–42*) involving e.g., the use of diffraction (*21, 43–46*). In this work, we harnessed the native dispersion properties of matter and trained a set of diffractive layers using deep learning to all-optically process a continuum of wavelengths in order to transform the spatial features of different object classes into a set of unique wavelengths, each representing one data class. This enabled us to use a single-pixel spectroscopic detector to perform optical classification of objects based on the spectral power encoded at these class-specific wavelengths. It should be emphasized that the task-specific spectral encoding provided through a trained diffractive optical network is a *single-shot* encoding for e.g., image classification, without the need for variable or structured illumination or dynamic spatial light modulators.

We demonstrated this novel machine vision framework by designing broadband diffractive optical networks that operate with pulsed illumination at terahertz wavelengths to achieve >96% blind testing accuracy for optical classification of handwritten digits (never seen by the network before) based on the spectral power at ten distinct wavelengths, each assigned to one digit/class. Using a plasmonic nanoantenna-based source and detector as part of a terahertz time-domain spectroscopy (THz-TDS) system (*47, 48*), and 3D-printed diffractive models, our experiments provided very good match to our numerical results, successfully inferring the classes/digits of the input objects by maximizing the power of the wavelength corresponding to the true label.

In addition to optical classification of objects through spectral encoding of data classes, we also demonstrate a shallow Artificial Neural Network (ANN) with two hidden layers that is successively trained (after the diffractive network training) to rapidly reconstruct the images of the classified objects based on their diffracted power spectra detected by a single-pixel spectroscopic detector. Using only 10 inputs, one for each class-specific wavelength, this shallow ANN is shown to successfully reconstruct images of the input objects even if they were incorrectly classified by the trained diffractive network. Considering the fact that each image of a handwritten digit is composed of 784 pixels, this shallow image reconstruction ANN, with an input vector size of 10, performs a form of image decompression to successfully decode the task-specific spectral encoding of the diffractive network (i.e., the optical front-end). Despite being a modest ANN with two hidden layers, the success of this task-specific image reconstruction network, i.e., the decoder, also emphasizes the vital role of the collaboration between a trainable optical front-end and an all-electronic ANN-based back-end (*39, 44*). In fact, our results also demonstrate that once the reconstructed images of the input objects that were initially misclassified by the diffractive network are fed back into the same network as new inputs, their optical classification is corrected, significantly improving the overall inference accuracy of the trained diffractive network.

We believe that the framework presented in this work would pave the way for the development of various new machine vision systems that utilize spectral encoding of object information to achieve a specific inference task in a resource-efficient manner, with low-latency, low power and low pixel count. These features are particularly important since large pixel-count of optical sensor arrays can put a burden on computational resources such as the allocated memory and the number of multiply-accumulate (MAC) units required for statistical inference or classification over a large image size; furthermore, such high resolution image sensors are not readily available at various parts of the electromagnetic spectrum, including for example far/mid-infrared and terahertz bands. The teachings of this work can also be extended to spectral domain interferometric measurement systems, such as Optical Coherence Tomography (OCT), Fourier Transform Infrared Spectroscopy (FTIR) and others, to create fundamentally new 3D imaging and sensing modalities integrated with spectrally encoded classification tasks performed through trained diffractive networks. While the presented approach utilized solely the native dispersion properties of matter, we also envision harnessing metamaterials and their engineered dispersion to design diffractive spectral encoding systems with additional degrees of freedom.

**Results**

Figure 1 illustrates our machine vision framework for spectral encoding of spatial information. A broadband diffractive network composed of layers is trained to transform the spatial information of the objects into the spectral domain through a pre-selected set of class-specific wavelengths measured by a single-pixel spectroscopic detector at the output plane; the resulting *spectral class scores* are denoted by the vector $\boldsymbol{s} = [s_0, s_1, \ldots, s_9]$ (Fig. 1a). Since in this work the learning task assigned to the diffractive network is the optical classification of the images of handwritten digits (MNIST database) (*49*), after its training and design phase, for a given input/test image it learns to channel relatively more power to the spectral component assigned to the correct class (e.g., digit '8' in Fig. 1a) compared to the other class scores; therefore, $\max(\boldsymbol{s})$ reveals the correct data class. As demonstrated in Fig. 1b, the same class score vector, $\boldsymbol{s}$, can also be used as an input to a shallow ANN with two hidden layers to reconstruct an image of the input object, decoding the spectral encoding performed by the broadband diffractive network.

Based on the system architecture shown in Fig. 1a, we trained broadband networks by taking the thickness of each pixel of a diffractive layer as a learnable variable (sampled at a lateral period of $\lambda_{min}/2$, where $\lambda_{min}$ refers to the smallest wavelength of the illumination bandwidth), and accordingly defined a training loss ($\mathcal{L}_D$) for a given diffractive network design:

$$\mathcal{L}_D = \mathcal{L}_I + \alpha \cdot \mathcal{L}_E + \beta \cdot \mathcal{L}_P \qquad (1),$$

where $\mathcal{L}_I$ and $\mathcal{L}_E$ refer to the loss terms related to the optical inference task (e.g., object classification) and the diffractive power efficiency at the output detector, respectively. The spatial purity loss, $\mathcal{L}_P$, on the other hand, has a rather unique aim of clearing the light intensity over a small region of interest surrounding the active area of the single-pixel detector to improve the robustness of the machine vision system for uncontrolled lateral displacements of the detector position with respect to the optical axis (see Supplementary Materials for detailed definitions of $\mathcal{L}_I$, $\mathcal{L}_E$ and $\mathcal{L}_P$). The hyperparameters, $\alpha$ and $\beta$, control the balance between the three major design factors, represented by these training loss terms.

To exemplify the performance of this design framework, using ten class-specific wavelengths uniformly distributed between $\lambda_{min}$ = 1.0 mm and $\lambda_{max}$ = 1.45 mm, a 3-layer diffractive optical network trained with $\alpha = \beta = 0$ achieves >96% blind testing accuracy for spectrally encoded optical classification of handwritten digits (see Table 1, 4[th] row). Fine tuning of the hyperparameters, $\alpha$ and $\beta$, yields broadband diffractive network designs that provide improved diffractive power efficiency at the output detector and partial insensitivity to misalignments without excessively sacrificing the inference accuracy. For example, using $\alpha = 0.03$ and $\beta = 0.1$, we have got 95.05% blind testing accuracy for spectrally encoded optical classification of handwritten digits with ~1% inference accuracy drop compared to the diffractive model trained with $\alpha = \beta = 0$, while at the same time achieving ~ 8-fold higher diffractive power efficiency at the output detector (see Table 1). Figure 2b illustrates the resulting layer thickness distributions of this diffractive network trained with $\alpha = 0.03$ and $\beta = 0.1$, setting a well-engineered example of the balance among inference accuracy, diffractive power efficiency at the output detector and misalignment resilience of the diffractive network.

Next, we fabricated these diffractive layers shown in Fig. 2b (trained with $\alpha = 0.03$, $\beta = 0.1$ to achieve 95.05% blind testing accuracy) together with 50 handwritten digits (5 per digit) randomly selected from the correctly-classified blind testing samples using 3D-printing (see Fig. 2a for the resulting diffractive network). Figure 2c also shows the THz-TDS setup with a plasmonic photoconductive detector that we used for the experimental validation of our machine vision framework (also see the Materials and Methods section). In this setup, the pulsed light emerging from a plasmonic photoconductive terahertz source is collimated and directed toward a square aperture with an area of 1 cm² (Fig. 2d), which serves as an entrance pupil to illuminate an unknown input object to be classified. As shown in Fig. 2d, we do not have any optical components or modulation layers between the illumination aperture and the object plane, indicating that there is *no* direct mapping between the spatial coordinates of the object plane and the spectral components of the illumination beam. Based on this experimental setup, the comparison between the power spectrum numerically generated using our trained forward model (dashed line) and its experimentally-measured counterpart (straight line) for 3 fabricated digits, as examples, is illustrated in Fig. 3a, providing a decent match between the two, also revealing the correct class inference in each case through $\max(s)$. Despite 3D-fabrication errors, possible misalignments and other sources of error in our setup, the match between the experimental and numerical testing of our diffractive network design was found to be 88% using 50 handwritten digits that were 3D-printed (see Fig. 3b).

For the same 3D-printed diffractive model (Fig. 2a, b), we also trained a shallow, fully-connected ANN with 2 hidden layers in order to reconstruct images of the unknown input objects based on the detected $s$. The training of this decoder ANN is based on the knowledge of (1) the class scores ($s = [s_0, s_1, ..., s_9]$) resulting from the trained diffractive network model, and (2) the corresponding input object images. Without any fine tuning of the network parameters for possible deviations between our numerical forward model and the experimental setup, when this shallow ANN was blindly tested on our experimental measurements ($s$), the reconstructions of the images of the handwritten digits were successful as illustrated in Fig. 1b (also see Fig. S6), further validating the presented framework as well as the experimental robustness of our diffractive network model (see Supplementary Materials for further details). It should be emphasized that this shallow ANN is trained to decode a highly compressed form of information that is spectrally encoded by a diffractive front-end and it uses only 10 numbers (i.e., $s_0, s_1, ..., s_9$) at its input to reconstruct an image that has >780 pixels. Stated differently this ANN performs a form of task-specific image decompression, the task being the reconstruction of the images of handwritten digits based on spectrally encoded inputs ($s$). In addition to performing task-specific image reconstruction, the presented machine vision framework can possibly be extended for the design of a *general-purpose* single-pixel imaging system based on spectral encoding, although here in this work we focused on the reconstruction of the classified object images (i.e., handwritten digits).

In addition to the diffractive network shown in Fig. 2 that achieved a numerical blind testing accuracy of 95.05%, we also 3D-fabricated and experimentally tested two additional diffractive network models to further evaluate the match between our numerical models and their experimental/physical counterparts. By using different $(\alpha, \beta)$ pairs for the loss function defined in Eq. (1), the balance between the optical inference accuracy and the two practical design merits, i.e., the diffractive power efficiency at the output detector and the insensitivity to misalignments, is shifted in these two new diffractive designs in favor of experimental robustness. Performance comparisons of these diffractive network models is summarized in Table 1 and Fig. 3c; for example, using $\alpha = 0.4$ and $\beta = 0.2$, the blind testing accuracy attained by the same 3-layer diffractive network architecture decreased to 84.02% for the handwritten digit classification task, while the diffractive power efficiency at the output detector increased by a factor of ~160 as well as the match between our experimental and numerical testing results increased to 96%. These results, as summarized in Fig. 3c and Table 1, further demonstrate the trade-off between the inference accuracy and the diffraction efficiency and experimental robustness of our diffractive network models.

To provide a mitigation strategy for this trade-off, next we introduced a collaboration framework between the diffractive network and its corresponding reconstruction ANN. This collaboration is based on the fact that our decoder ANN can faithfully reconstruct the images of the input objects using the spectral encoding present in $s$, even if the optical classification is incorrect, pointing to a wrong class through max($s$). We observed that by feeding the decoder ANN's reconstructed images back to the diffractive network as *new* inputs we can have it correct its initial wrong inference (see Fig. 4 and Fig. S2). Through this collaboration between the diffractive network and its decoder ANN, we improved the overall inference accuracy of a given diffractive network model as summarized in Fig. 3c and Table 1. For example, for the same, highly-efficient diffractive network model that was trained using $\alpha = 0.4$ and $\beta = 0.2$, the blind testing accuracy for handwritten digit classification increased from 84.02% to 91.29% (see Figs. 3c and 5b), demonstrating a substantial improvement through the collaboration between the decoder ANN and the broadband diffractive network. A close examination of Fig. 5 and the provided confusion matrices reveals that the decoder ANN, through its image reconstruction, helped to correct 870 misclassifications of the diffractive network, resulting in an overall gain/improvement of 7.27% in the blind inference performance of the optical network. Similar analyses for the other diffractive network models are also presented in Figs. S3, S4 and S5 of Supplementary Materials.

In this collaboration between the diffractive network and its corresponding shallow decoder, the training loss function of the latter (ANN) was coupled to the classification performance of the diffractive network. In other words, in addition to a structural loss function ($\mathcal{L}_S$) that is needed for a high-fidelity image reconstruction, we also added a second loss term that penalized the ANN by a certain weight if its reconstructed image cannot be correctly classified by the diffractive network (see the Supplementary Materials). This ensures that the collaboration between the optical encoder and its corresponding decoder ANN is constructive, i.e., the overall classification accuracy is improved through the feedback of the reconstructed images onto the diffractive network as new inputs. Based on this collaboration scheme, the general loss function of the decoder ANN can be expressed as:

$$\mathcal{L}_{\text{Recon}} = \gamma \cdot \mathcal{L}_S(\boldsymbol{O}_{\text{recon}}, \boldsymbol{O}_{\text{input}}) + (1-\gamma) \cdot \mathcal{L}_I \qquad (2),$$

where $\mathcal{L}_S$ refers to a structural loss term, e.g. Mean Absolute Error (MAE) or reversed Huber ("BerHu") loss (*50*, *51*), computed through pixel-wise comparison of the reconstructed image ($\boldsymbol{O}_{\text{recon}}$) with the ground truth object image ($\boldsymbol{O}_{\text{input}}$) (see Supplementary Materials for details). The second term in Eq. (2), $\mathcal{L}_I$, refers to the same loss function used in the training of the diffractive network (front-end) as in Eq. (1), except this time it is computed over the new class scores, $\boldsymbol{s}'$, obtained by feeding the reconstructed image, $\boldsymbol{O}_{\text{recon}}$, back to the same diffractive network (see Fig. 5 and Fig. S1). Eq. (2) is *only* concerned with the training of the image reconstruction ANN, and therefore, the parameters of the decoder ANN are updated through standard error backpropagation, while the diffractive network model is preserved.

Table 1 summarizes the performance comparison of different loss functions employed to train the decoder ANN and their impact on the improvement of the classification performance of the diffractive network. Compared to the case when $\gamma = 1$, which refers to independent training of the reconstruction ANN without taking into account $\mathcal{L}_I$, we see significant improvements in the inference accuracy of the diffractive network through $\max(\boldsymbol{s}')$ *when* the ANN has been penalized during its training (with e.g., $\gamma = 0.95$) if its reconstructed images cannot be correctly classified by the diffractive network (refer to the Supplementary Materials for details). Stated differently, the use of $\mathcal{L}_I$ term in Eq. (2) for the training of the decoder ANN tailors the image reconstruction space to generate object features that are more favorable for the diffractive optical classification, while also retaining its reconstruction fidelity to the ground truth object, $\boldsymbol{O}_{\text{input}}$, by the courtesy of the structural loss term, $\mathcal{L}_S$, in Eq. (2).

The performance of the presented spectral encoding-based machine vision framework can be further improved using a differential class encoding strategy (*45*). For this aim, we explored the use of two different wavelengths to encode each class score: instead of using 10 discrete wavelengths to represent a spectral class score vector, $\boldsymbol{s} = [s_0, s_1, ..., s_9]$, we considered encoding the spatial information of an object into 20 different wavelengths ($s_{0+}, s_{0-}, s_{1+}, s_{1-}, ..., s_{9+}, s_{9-}$) that are paired in groups of two in order to *differentially* represent each spectral class score, i.e., $\Delta s_c = \frac{s_{c,+} - s_{c,-}}{s_{c,+} + s_{c,-}}$. In this *differential spectral encoding* strategy, the trained diffractive network makes an inference based on max($\Delta s$) resulting from the spectral output at the single-pixel detector. With this spectrally encoded differential classification scheme, we numerically attained 96.82% optical classification accuracy for handwritten digits (see Table 1 and Fig. S8).

As an alternative to the shallow decoder ANN with 2-hidden layers used earlier, we also explored the use of a much deeper ANN architecture (*52*) as the image reconstruction network in our spectrally encoded machine vision framework. For this, the output of the 2-hidden layer fully-connected network (with an input of $s$) is further processed by a U-Net-like deep convolutional ANN with skip connections and a total of >1.4M trainable parameters in order to reconstruct the images of handwritten digits using $s$. We found out that the collaboration of the diffractive networks with this deeper ANN architecture yielded only marginal improvements over the classification accuracies presented in Table 1. For example, when the diffractive optical network design shown in Fig. 2b ($\alpha = 0.03, \beta = 0.1$) was paired with this deep decoder ANN (through the feedback depicted in Fig. 4), the blind classification accuracy increased to 95.52% compared to the 95.37% provided by the shallow decoder ANN with 2-hidden layers. As another example, for the diffractive optical network trained with $\alpha = 0.4$ and $\beta = 0.2$, the collaboration with the deep convolutional ANN provides a classification accuracy of 91.49%, which is a minor improvement with respect to the 91.29% accuracy produced through the shallow ANN, falling short to justify the disadvantages of using a deeper ANN-based decoder architecture in terms of its slower inference speed and more power consumption per image reconstruction.

The function of the decoder ANN, up to this point, has been to reconstruct the images of the unknown input objects based on the encoding present in the spectral class scores, $s$. Therefore, it is important to emphasize that, in these earlier results, the classification is performed optically, through the spectral output of the diffractive network, and the function of the decoder ANN is *not* to infer a data class, but rather to reconstruct the image of the object that is classified by the diffractive optical network using max($s$). As an alternative strategy, we also explored making use of the decoder ANN for a different task: to directly classify the objects based on the spectral encoding ($s$) provided by the diffractive network. In this case, the decoder ANN is solely focused on improving the classification performance with respect to the optical inference results that are achieved using max($s$). For example, based on the spectral class scores encoded by the diffractive optical networks that achieved 95.05% and 96.07% blind testing accuracy for handwritten digit classification using max($s$), a fully-connected, shallow *classification* ANN with 2-hidden layers improved the classification accuracy to 95.74 and 96.50%, respectively. Compared to the accuracy values presented in Table 1, these numbers indicate that a slightly better classification performance is possible, provided that the image reconstruction is not essential for the target application, and can be replaced with a shallow *classification* decoder ANN that takes $s$ as its input.

In the earlier machine vision systems that we have presented so far, the diffractive optical network and the corresponding back-end ANN have been separately trained, i.e., after the training of the diffractive network for optical image classification, the back-end ANN was trained based on the spectral encoding of the converged diffractive network model, yielding either the reconstruction ANN or the classification ANN, as discussed earlier. As an alternative strategy, such hybrid systems can also be *jointly-trained*, through the error backpropagation between the electronic ANN and the diffractive optical front-end (*44, 53*). Here we demonstrated this opportunity using the MNIST dataset and jointly-trained a diffractive network with an image *reconstruction* ANN at the back-end; in the next paragraphs the same approach will also be extended to jointly-train a diffractive network with a *classification* ANN at the back-end, covering a different dataset (EMNIST) (*54*). In our joint-training of hybrid network systems composed of a diffractive network and a reconstruction ANN, we used a linear superposition of two different loss functions to optimize both the optical classification accuracy and the image reconstruction fidelity; see Eq. S22 in Supplementary Materials and Supplementary Table S2. Through this linear superposition, we explored the impact of different relative weights of these loss functions on (1) the image classification accuracy of the diffractive network, and (2) the quality of the image reconstruction performed by the back-end ANN. For this goal, we changed the relative weight ($\xi$) of the optical classification loss term in order to shift the attention of the hybrid design between these two tasks. For instance, when the weight of the optical classification loss is set to be zero ($\xi = 0$), the entire hybrid system becomes a computational single-pixel imager that ignores the optical classification accuracy and focuses solely on the image reconstruction quality; as confirmed in Supplementary Figures S11, S12 and Supplementary Table S2, this choice ($\xi = 0$) results in a substantial degradation of the optical image classification accuracy with a considerable gain in the image reconstruction fidelity, as expected. By using different relative weights, one can achieve a sweet spot in the joint-training of the hybrid network system, where both the optical image classification accuracy and the ANN image reconstruction fidelity are very good; see e.g., $\xi = 0.5$ in Supplementary Table S2, Figs. S11 and S12.

We also investigated the inference performance of these hybrid systems in terms of the number of wavelengths that are simultaneously processed through the diffractive network. For this, we jointly-trained hybrid systems that assign a group of wavelengths to each data class; inference of an object class is then based on the *maximum* average power accumulated in these selected spectral bands, where each band represents one data class (see Supplementary Materials for further details). Our results, summarized in Supplementary Table S2, reveal that assigning e.g., 5 distinct wavelengths to each data class (i.e., a total of 50 wavelengths for 10 data classes), achieved a similar optical classification accuracy, compared to their counterparts that encoded the objects' spatial information using fewer wavelengths. This indicates that the diffractive networks can be designed to simultaneously process a larger number of wavelengths to successfully encode the spatial information of the input FOV into spectral features.

To further explore the capabilities of the presented single-pixel spectroscopic machine vision framework for more challenging image classification tasks beyond handwritten digits, we utilized the EMNIST dataset (*54*), containing 26 object classes, corresponding to handwritten capital letters (see Figure S13). For this EMNIST image dataset, we trained non-differential and differential diffractive classification networks, encoding the information of the object data classes into the output power of 26 and 52 distinct wavelengths, respectively. Furthermore, to better highlight the benefits of the collaboration between the optical and electronic networks, we also *jointly-trained* hybrid network systems that use a shallow *classification* ANN (with 2 hidden layers) described earlier to extract the object class from the spectral encoding performed by the diffractive optical front-end, through a single-pixel detector, same as before. Supplementary Table S1 summarizes our results on this 26-class handwritten capital letter image dataset. First, a comparison between the all-optical diffractive classification networks and the jointly-trained hybrid network systems highlights the importance of the collaboration between the optical and electronic networks: the *jointly-trained* hybrid systems (where a diffractive network is followed by a *classification encoder* ANN) can achieve higher object classification accuracies (see Supplementary Table S1). For example, a jointly-trained hybrid network using 52 encoding wavelengths that are processed through 3 diffractive layers and a shallow decoder ANN achieved a classification accuracy of 87.68% for EMNIST test dataset, which is >2% higher compared to the inference accuracy attained solely by an optical diffractive network design based on differential spectral encoding using the same 52 wavelengths (Supplementary Table S1). The results presented in Supplementary Table S1 further reveal that both the jointly-trained hybrid systems and the optical diffractive classification systems that utilize 52 distinct wavelengths to encode the spatial information of the objects achieve higher classification accuracies compared to their counterparts that are designed to process 26 wavelengths.

**Discussion**

Even though Eq. (1) tries to find a balance among the optical inference accuracy, detector photon efficiency and resilience to possible detector misalignment, there are other sources of experimental errors that contribute to the physical implementations of trained diffractive networks. First, due to multi-layer layout of these diffractive networks, any inter-layer misalignments might have contributed to some of the errors that we observed during the experiments. In addition, our optical forward model does not take into account multiple reflections that occur through the diffractive layers. These are relatively weaker effects that can be mitigated by e.g., time-gating of the detector output and/or using anti-reflection coatings that are widely employed in the fabrication of conventional optical components. Moreover, measurement errors that might have taken place during the characterization of the dispersion of the diffractive-layer material can cause our numerical models to slightly deviate from their physical implementations. Furthermore, 3D fabrication errors stemming from printing overflow and crosstalk between diffractive features on the layers can also contribute to some of the differences observed between our numerical and experimental results. Supplementary Fig. S14 illustrates a comparison between one of the 3D-printed diffractive layers and its numerical design, exemplifying some of these fabrication-related imperfections that are experimentally observed.

The negative effects of some of these experimental errors outlined above can be mitigated by modelling undesired physical system variations over random variables that are incorporated as part of the optical forward model used in the deep learning-based training of the diffractive network (*53,55*). Thereby, the evolution of the parameter space of the underlying diffractive layers can be regulated to preserve their collective inference accuracy despite e.g., misalignments. We explored the impact of this idea for building misalignment resilience in jointly-trained, hybrid MNIST classification networks formed by a 3-layer spectral encoder diffractive front-end and a shallow *classification* ANN (with 2 hidden layers). Based on this testbed, we modeled the lateral misalignments of the spectral encoder diffractive layers by defining two independent, uniformly distributed random variables per layer, $D_x^l \sim U(-\Delta_x^l, \Delta_x^l)$ and $D_y^l \sim U(-\Delta_y^l, \Delta_y^l)$, representing the displacement of layer $l$ with respect to its ideal location in $x$ and $y$ directions, respectively. The hyperparameters, $\Delta_x^l$ and $\Delta_y^l$, determine the range of the positioning error along the corresponding axis. During the training (*tr*) phase, $D_x^l$ and $D_y^l$, were randomly updated at each iteration, uniformly taking random values from the range set by $\Delta_x^l = \Delta_y^l = \Delta_{tr}^l$. Such a training strategy (which we term as vaccination) introduces new perturbed diffractive layer locations at each update step and, as a result, guides the evolution of the spectral encoder diffractive layers to a solution that is more resilient against misalignments, allowing the diffractive networks to maintain their optical inference accuracy over larger margins of physical misalignments.

To demonstrate the impact of the outlined training strategy, we quantified the blind inference accuracies achieved by various vaccinated and nonvaccinated diffractive single-pixel machine vision systems under a series of misalignments as shown in Supplementary Fig. S15. In Supplementary Fig. S15a, we report the vaccination results when only the middle diffractive layer ($l = 2$) is misaligned from its ideal location, meaning that the centers of the 1$^{st}$ and 3$^{rd}$ diffractive layers coincide with the optical axis, i.e., $\Delta_x^1 = \Delta_y^1 = \Delta_{test}^1 = \Delta_{tr}^1 = 0.0$ and $\Delta_x^3 = \Delta_y^3 = \Delta_{test}^3 = \Delta_{tr}^3 = 0.0$. In Supplementary Fig. S15b, on the other hand, all three diffractive layers experience random lateral shifts along both $x$ and $y$ axes during the testing phase. In our analyses, we also investigated the effect of the single-pixel detector size on the misalignment resilience of these hybrid neural network systems and accordingly trained vaccinated and nonvaccinated single-pixel spectral encoder diffractive systems, each with a detector active area of 2×2 mm$^2$, 4×4 mm$^2$ and 8×8 mm$^2$. As depicted in Supplementary Figs. S15a,b, almost independent of the active area of the single-pixel detector, the classification accuracy of the *nonvaccinated* hybrid networks (blue) are rather sensitive to mechanical misalignments of the diffractive network layers. The vaccinated networks, on the other hand, can maintain their blind inference accuracy over a wider range of misalignments, which is confirmed in both Supplementary Fig. S15a and Fig. S15b. Furthermore, a comparison of the classification accuracies provided by the vaccinated hybrid network systems reveals that the design with a larger active area (8×8 mm$^2$) single-pixel detector achieves better resilience over misalignments (see Supplementary Fig. S15b).

Without loss of generality, in this work we used a 3-layer diffractive network architecture to encode the spatial features of the object field-of-view into the output power spectrum for single-pixel machine vision. It is important to note that if the material absorption of the diffractive layers is lower and/or the signal-to-noise ratio of the single-pixel detector is increased, the optical inference accuracy of the presented network designs could be further improved by e.g., increasing the number of diffractive layers or the number of learnable features (i.e., neurons) within the optical network (*44*, *56*). Compared to using wider diffractive layers, increasing the number of diffractive layers offers a more practical method to enhance the information processing capacity of diffractive networks, since training higher numerical aperture diffractive systems through image data is in general relatively harder (*56*). Despite their improved generalization capability, such deeper diffractive systems composed of larger numbers of diffractive layers would partially suffer from increased material absorption and surface back-reflections. However, one should note that the optical power efficiency of a broadband network also depends on the size of the output detector. For example, the relatively lower power efficiency numbers reported in Table 1 are by and large due to the small size of the output detector used in these designs ($2 \times \lambda_{\min}$) and can be substantially improved by using a detector with a much larger active area (*57*).

In conclusion, we demonstrated an optical machine vision system composed of trained diffractive layers to encode the spatial information of objects into the power spectrum of diffracted light, which is used to perform optical classification of unknown objects with a single-pixel spectroscopic detector. We also showed that shallow, low-complexity ANNs can be used as decoders to reconstruct images of the input objects based on the spectrally-encoded class scores, demonstrating task-specific image decompression. Although we used terahertz pulses to experimentally validate our spectrally encoded machine vision framework, it can be broadly adopted for various applications covering other parts of the electromagnetic spectrum. In addition to object recognition, this machine vision concept can also be extended to perform other learning tasks such as scene segmentation, multi-label classification, as well as to design single or few pixel, low-latency super-resolution imaging systems by harnessing the spectral encoding provided by diffractive networks coupled with shallow decoder ANNs. We also envision that dispersion engineered material systems such as metamaterials will open up a new design space for enhancing the inference and generalization performance of spectral encoding through diffractive optical networks. Finally, the methods presented in this work would create new 3D imaging and sensing modalities that are integrated with optical inference and spectral encoding capabilities of broadband diffractive networks, and can be merged with some of the existing spectroscopic measurement techniques, such as OCT, FTIR and others, to find various new applications in biomedical imaging, analytical chemistry, material science and other fields.

## Materials and Methods

**Terahertz time-domain spectroscopy setup.** The schematic diagram of the terahertz time-domain spectroscopy (THz-TDS) setup is shown in Fig. 2d. We employed a Ti:Sapphire laser (Coherent MIRA-HP) in a mode-locked operation mode to generate femtosecond optical pulses at a center wavelength of 780 nm. The laser beam was first split in two parts. One part of the beam illuminated the terahertz source, a plasmonic photoconductive nano-antenna array *(48)*, to generate terahertz pulses. The other part of the laser beam passed through an optical delay line and illuminated the terahertz detector, which was another plasmonic photoconductive nano-antenna array offering high-sensitivity and broadband operation *(47)*. The generated terahertz radiation was collimated and guided to the terahertz detector using an off-axis parabolic mirror. The output signal as a function of the delay line position, which provides the temporal profile of the detected terahertz pulses, was amplified using a current pre-amplifier (Femto DHPCA-100) and detected with a lock-in amplifier (Zurich Instruments MFLI). For each measurement, 10 time-domain traces were captured over 5 s and averaged. The acquired time-domain signal has a temporal span of 400 ps and its power spectrum was obtained through a Fourier transform. Overall, the THz-TDS system offers signal-to-noise ratio levels of >90 dB and observable bandwidths exceeding 5 THz.

The 3D-printed diffractive optical network was placed between the terahertz source and the detector. It consisted of an input aperture, an input object, three diffractive layers and an output aperture, as shown in Fig. 2d, with their dimensions and spacing annotated. Upon their training in a computer, the diffractive optical networks were fabricated using a 3D printer (Objet30 Pro, Stratasys Ltd.) with a UV curable material (VeroBlackPlus RGD875, Stratasys Ltd.). A $1\times 1$ cm square aperture was positioned at the input plane serving as an entrance pupil for the subsequent optical system. The terahertz detector has an integrated Si lens in the form of a hemisphere directly attached to the backside of the chip. This Si lens was modelled as an achromatic flat Si slab with a thickness of 0.5 cm and a refractive index of 3.4 in our optical forward model. During the experiments, a $2\times 2$ mm output aperture was placed at the output plane, right before the terahertz detector, to shrink the effective area of the Si lens, ensuring that the uniform slab model assumed during the training forward model accurately translates into our experimental setup. The input and output apertures as well as the 3D-printed objects were coated with aluminum to block terahertz radiation outside the transparent openings and object features. Furthermore, a 3D-printed holder (Fig. 2a) was designed to support and align all of the components of the diffractive setup.

**Supplementary Materials:** This file contains Supplementary Methods, Supplementary Figures S1-S16 and Supplementary Tables S1-S2.

# Figures

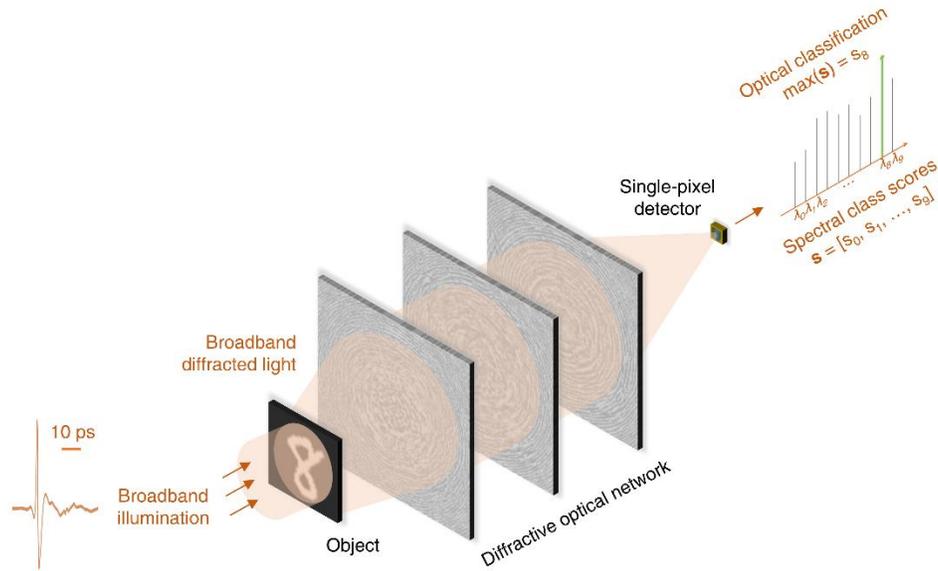

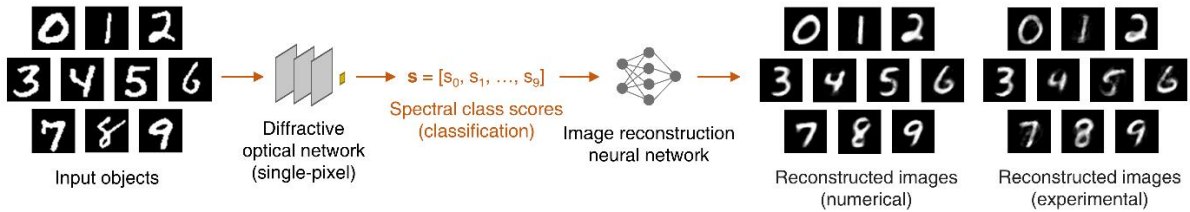

**Figure 1. Schematics of spectral encoding of spatial information for object classification and image reconstruction. a**, Optical layout of the single detector machine vision concept for spectrally-encoded classification of objects, e.g., the images of handwritten digits. As an example, a handwritten digit '8' is illuminated with a broadband pulsed light, and the subsequent diffractive optical network transforms the object information into the power spectrum of the diffracted light collected by a single detector. The object class is determined by the maximum of the spectral class scores, $s$, defined over a set of discrete wavelengths, each representing a data class (i.e., digit). **b**, Schematic of task-specific image reconstruction using the diffractive network's spectral class scores as input. A separately-trained shallow ANN (with 2-hidden layers) recovers the images of handwritten digits from the spectral information encoded in $s$. Each reconstructed image is composed of >780 pixels, whereas the input vector, $s$, has 10 values.

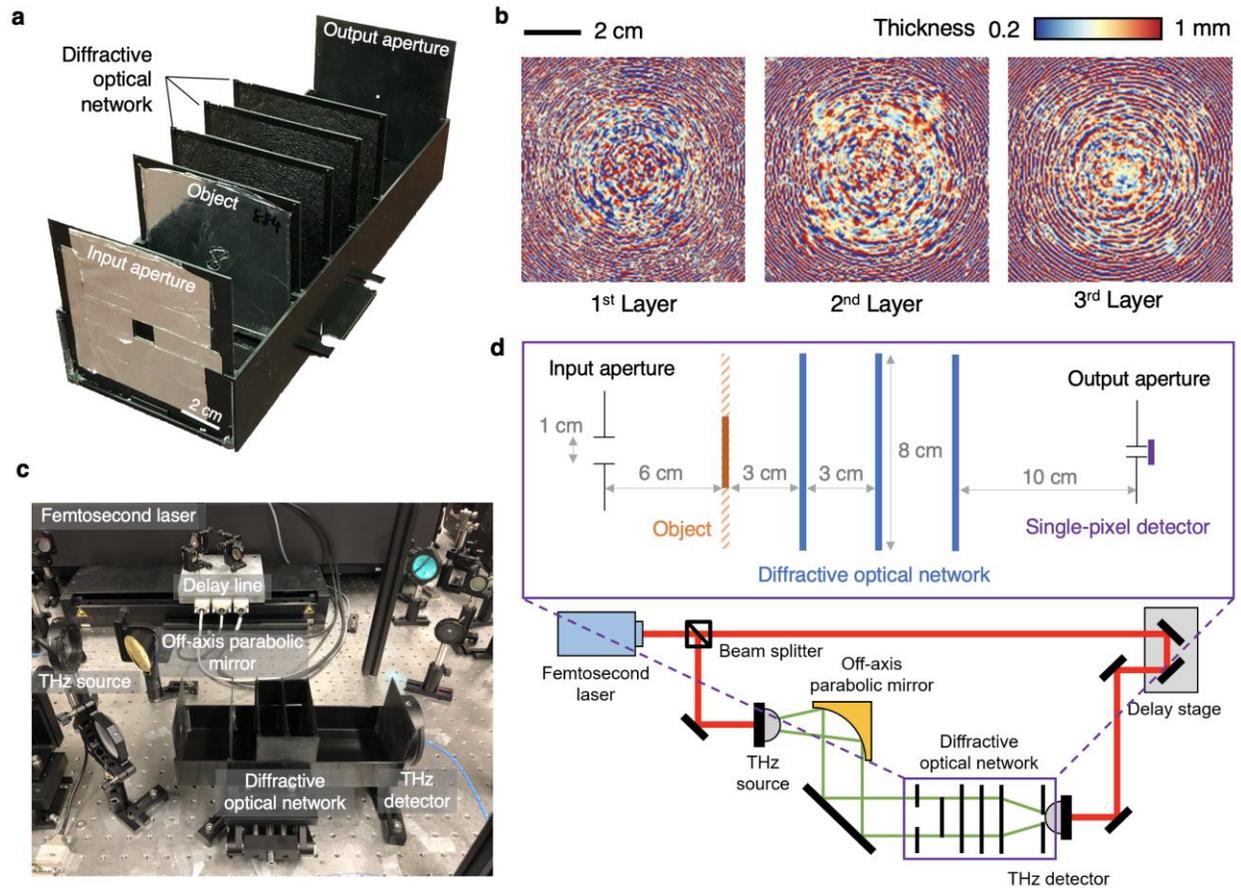

**Figure 2. Experimental setup. a**, A 3D-printed diffractive network. **b**, Learned thickness profiles of the three diffractive layers in (**a**). **c**, Photograph of the experimental setup. **d**, **Top:** physical layout of the diffractive optical network setup, zoomed-in version of the bottom part. The object is a binary handwritten digit (from MNIST data), where the opaque regions are coated with aluminum to block the light transmission. **Bottom:** schematic of the THz-TDS setup. Red lines depict the optical path of the femtosecond pulses generated by a Ti:Sapphire laser operating at 780 nm wavelength. Green lines indicate the optical path of the terahertz pulse (peak frequency ~ 500 GHz, observable bandwidth ~ 5 THz), which is modulated by the 3D-printed diffractive neural network to spectrally encode the task-specific spatial information of the objects.

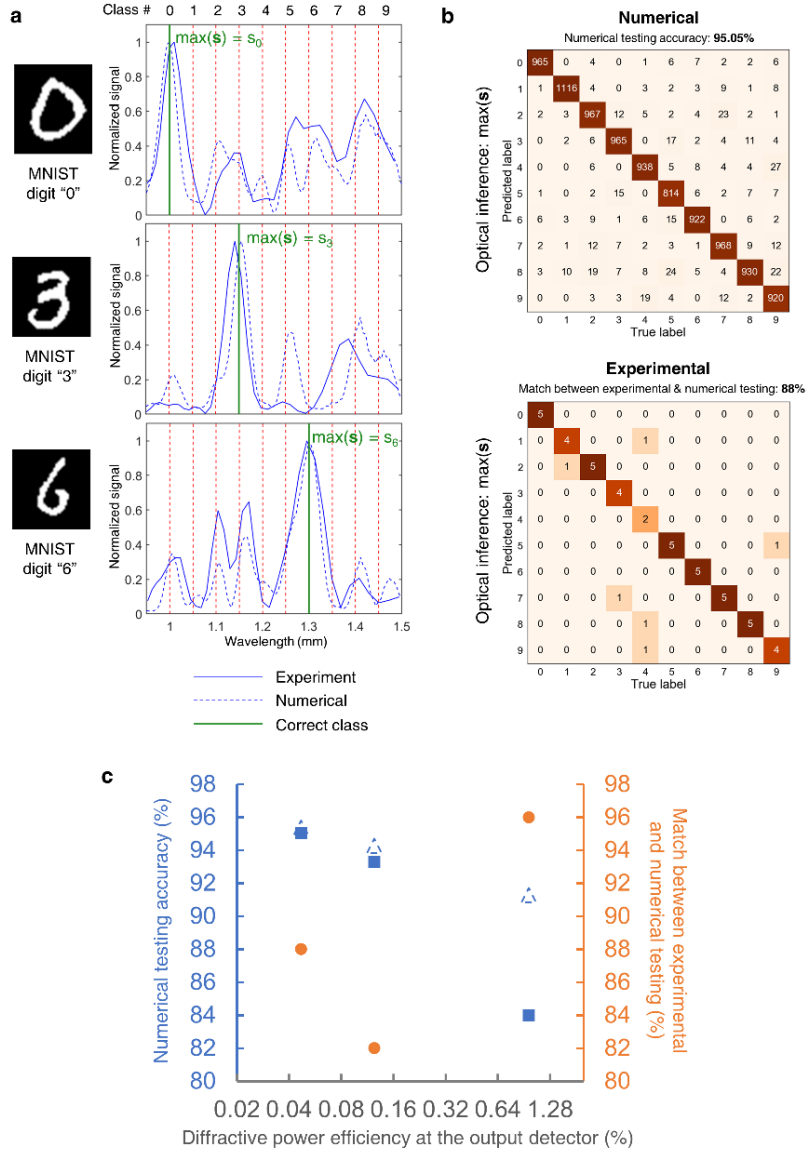

**Figure 3. Spectrally encoded optical classification of handwritten digits with a single detector. a**, Experimentally measured (blue-solid line) and the numerically computed (blue-dashed line) output power spectra for optical classification of three different handwritten digits, shown as examples. The object class is determined by the maximum of the spectral class scores, ***s***, defined over a set of discrete wavelengths, each representing a digit. **b**, **Top:** confusion matrix, summarizing the numerical classification performance of the diffractive optical network that attains a classification accuracy of 95.05% over 10,000 handwritten digits in the blind testing set. **Bottom:** confusion matrix for the experimental results obtained by 3D-printing of 50 handwritten digits randomly selected from the numerically successful classification samples in the blind testing set. An 88% match between the experimentally inferred and the numerically computed object classes is observed. **c**, Comparison of 3 different diffractive networks that were trained, fabricated and experimentally tested in terms of (1) their numerical blind testing accuracies (blue solid squares), (2) the match between experimentally measured and numerically predicted object classes (orange solid circles), and (3) the inference accuracy achieved by feeding the decoder ANN's reconstructed images back to the diffractive network as new inputs (blue dashed triangles).

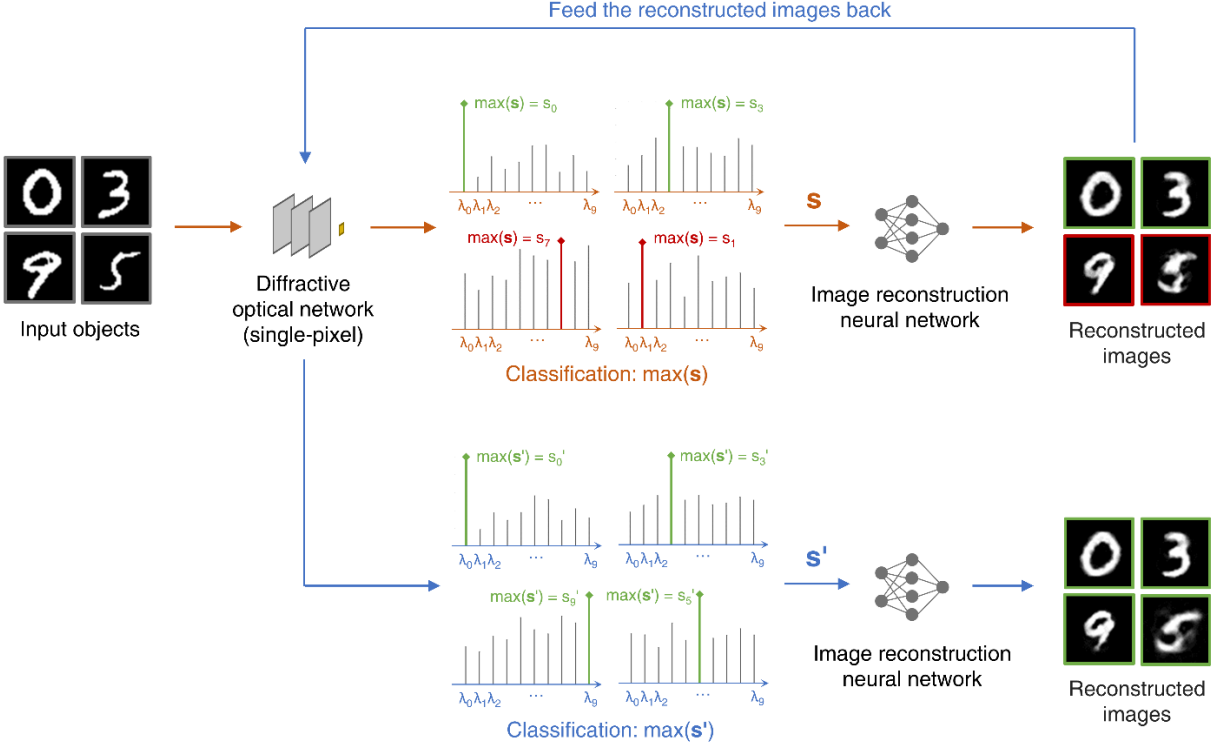

**Figure 4. Illustration of the coupling between the image reconstruction ANN and the diffractive network.** Four MNIST images of handwritten digits are used here for illustration of the concept. Two of the four samples, "0" and "3", are correctly classified by the diffractive network based on max($s$) (top green lines), while the other two, "9" and "5", are misclassified as "7" and "1", respectively (top red lines). Using the same class scores ($s$) at the output detector of the diffractive network, a shallow decoder ANN digitally reconstructs the images of the input objects. Next, these images are cycled back to the diffractive optical network as new input images and the new spectral class scores $s'$ are inferred accordingly, where all of the four digits are correctly classified through max($s'$) (bottom green lines). Finally, these new spectral class scores $s'$ are used to reconstruct the objects again using the same image reconstruction ANN. The blind testing accuracy of this diffractive network for handwritten digit classification increased from 84.02% to 91.29% using this feedback loop (see Fig. 3c and Fig. 5b). This image reconstruction decoder ANN was trained using the MAE loss and softmax-cross-entropy loss (see Eq. 2).

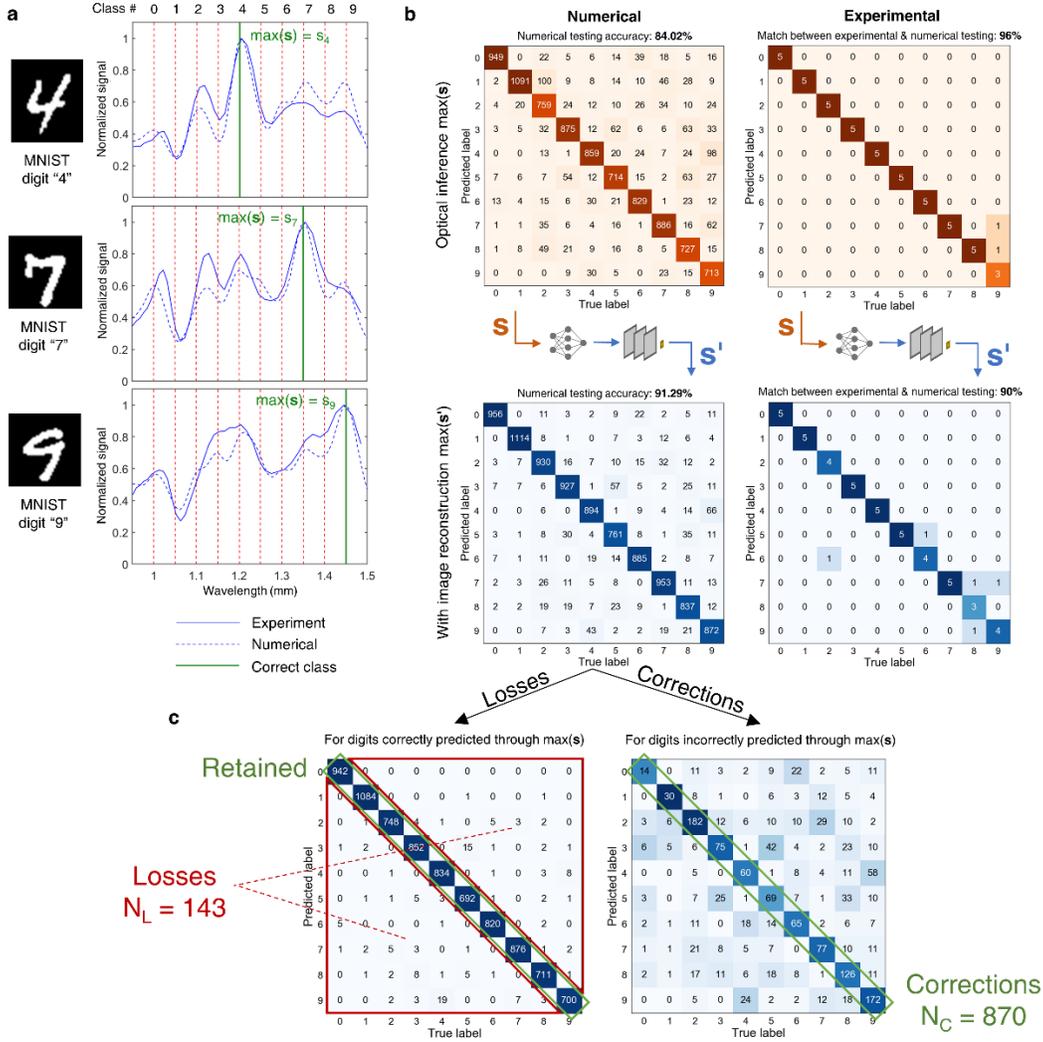

**Figure 5. Blind testing performance of an efficient diffractive network and its coupling with a corresponding decoder ANN.**
**a**, Experimentally measured (blue-solid line) and the numerically computed (blue-dashed line) output power spectra for optical classification of three different handwritten digits, shown as examples. **b**, **Top Left:** confusion matrix summarizing the numerical classification performance of the diffractive network that attains a classification accuracy of 84.02% over 10,000 handwritten digits in the blind testing set. **Top Right:** confusion matrix for the experimental results obtained by 3D-printing 50 handwritten digits randomly selected from the numerically successful classification samples in the blind testing set. A 96% match between the experimentally inferred and the numerically computed object classes is observed. **Bottom Left:** confusion matrix provided by $\max(s')$, computed by feeding the reconstructed images back to the diffractive network. A blind testing accuracy of 91.29% is achieved, demonstrating a significant classification accuracy improvement of 7.27% (also see Fig. 4). **Bottom Right:** confusion matrix for the experimental results using the same 50 digits. **c, Left:** same as the bottom left matrix in (**b**), but solely for the digits that are correctly predicted by the optical network. Its diagonal entries can be interpreted as the digits that are retained to be correctly predicted, while its off-diagonal entries represent the "losses" after the image reconstruction and feedback process. **Right:** same as the left one, but solely for the digits that are incorrectly classified by the optical network. Its diagonal entries indicate the optical classification "corrections" after the image reconstruction and feedback process. The number $N_C - N_L = 727$ is the classification accuracy "gain" achieved through $\max(s')$, corresponding to a 7.27% increase in the numerical testing accuracy of the diffractive model (also see Fig. 3c).

## Tables

| Diffractive network | Diffractive power efficiency at the output detector: $\eta$ (%) | Testing accuracy $\max(s)$ (%) | Testing accuracy $\max(s')$ (%) |
|---|---|---|---|
| 10 wavelengths, $\alpha = 0.4$, $\beta = 0.2$ (Fig. 5) $\mathbf{s} = [s_0, s_1, \ldots, s_9]$ | $0.966 \pm 0.465$ | 84.02 | MAE: 84.03<br>MAE + SCE: 91.29<br>BerHu + SCE: 91.06 |
| 10 wavelengths, $\alpha = 0.08$, $\beta = 0.2$ (Fig. S4) $\mathbf{s} = [s_0, s_1, \ldots, s_9]$ | $0.125 \pm 0.065$ | 93.28 | MAE: 91.31<br>MAE + SCE: 94.27<br>BerHu + SCE: 94.02 |
| 10 wavelengths, $\alpha = 0.03$, $\beta = 0.1$ (Fig. 3, Fig. S3) $\mathbf{s} = [s_0, s_1, \ldots, s_9]$ | $0.048 \pm 0.027$ | 95.05 | MAE: 93.40<br>MAE + SCE: 95.32<br>BerHu + SCE: 95.37 |
| 10 wavelengths, $\alpha = \beta = 0$ (Fig. S5) $\mathbf{s} = [s_0, s_1, \ldots, s_9]$ | $0.006 \pm 0.004$ | 96.07 | MAE: 94.58<br>MAE + SCE: 96.26<br>BerHu + SCE: 96.30 |
| 20 wavelengths (Differential), $\alpha = \beta = 0$ (Fig. S8) $\mathbf{s_D} = [s_{0+}, s_{0-}, s_{1+}, s_{1-}, \ldots, s_{9+}, s_{9-}]$ $\mathbf{s} = \Delta\mathbf{s} = [\Delta s_0, \Delta s_1, \ldots, \Delta s_9]$ | $0.004 \pm 0.002$ | 96.82 | MAE: 90.15<br>MAE + SCE: 96.81<br>BerHu + SCE: 96.64 |

**Table 1. Numerical blind testing accuracies of different diffractive network models and their integration with decoder image reconstruction ANNs.** The diffractive optical networks presented in the first 3 rows were trained with different $(\alpha, \beta)$ pairs for experimental validation, resulting in different diffractive power efficiencies at the output detector, while the model in the 4th and 5th row was trained with $\alpha = \beta = 0$. The mean diffractive power efficiencies ($\eta$) of the diffractive network models were calculated at the output detector, considering the whole testing dataset, represented with the corresponding standard deviations (see Supplementary Materials for details).